\documentclass[11pt,onecolumn]{article}         
\usepackage[top=3.6cm, bottom=3.2cm, left=2.3cm, right=2.3cm]{geometry}

\usepackage{color}
\usepackage{graphicx}
\usepackage{cite}
\usepackage[cmex10]{amsmath}
\usepackage{amssymb}
\usepackage{subfigure}
\usepackage{rotating}
\usepackage{multirow}
\usepackage{bm}
\usepackage[colorlinks=true,linkcolor=red,citecolor=blue]{hyperref}
\newcommand{\ie}{\emph{i.e.}}

\begin{document}

\title{Learning RoI Transformer for Detecting \\ Oriented Objects in Aerial Images}

{
\author{Jian Ding, Nan Xue, Yang Long, Gui-Song Xia\thanks{Corresponding author: guisong.xia@whu.edu.cn.}, Qikai Lu\\
{\em LIESMARS-CAPTAIN, Wuhan University, Wuhan, 430079, China}\\
{\tt\small \{jian.ding, xuenan, longyang, guisong.xia, qikai\_lu\}@whu.edu.cn}
}
}

\maketitle

\begin{abstract}
Object detection in aerial images is an active yet challenging task in computer vision because of the birdview perspective, the highly complex backgrounds, and the variant appearances of objects. Especially when detecting densely packed objects in aerial images, methods relying on horizontal proposals for common object detection often introduce mismatches between the Region of Interests (RoIs) and objects. 
This leads to the common misalignment between the final object classification confidence and localization accuracy. 
Although rotated anchors have been used to tackle this problem, the design of them always multiplies the number of anchors and dramatically increases the computational complexity.  
In this paper, we propose a {\bf RoI Transformer} to address these problems. More precisely, to improve the quality of region proposals, we first designed a Rotated RoI (RRoI) learner to transform a Horizontal Region of Interest (HRoI) into a Rotated Region of Interest (RRoI). Based on the RRoIs, we then proposed a Rotated Position Sensitive RoI Align (RPS-RoI-Align) module to extract rotation-invariant features from them for boosting subsequent classification and regression. 
Our RoI Transformer is with light weight and can be easily embedded into detectors for oriented object detection. A simple implementation of the RoI Transformer has achieved state-of-the-art performances on two common and challenging aerial datasets, \ie, DOTA and HRSC2016, with a neglectable reduction to detection speed. Our RoI Transformer exceeds the deformable Position Sensitive RoI pooling when oriented bounding-box annotations are available.
Extensive experiments have also validated the flexibility and effectiveness of our RoI Transformer. The results demonstrate that it can be easily integrated with other detector architectures and significantly improve the performances. 
\end{abstract}

\section{Introduction}
Object detection in aerial images aims at locating objects of interest (e.g., vehicles, airplanes) on the ground and identifying their categories. With more and more aerial images being available, object detection in aerial images has been a specific but active topic in computer vision~\cite{roatation-invariant-cvpr,long2017accurate,rotation-invariant, deng2017toward}. However, unlike natural images that are often taken from horizontal perspectives, aerial images are typically taken with birdviews, which implies that objects in aerial images are always arbitrary oriented.
Moreover, the highly complex background and variant appearances of objects further increase the difficulty of object detection in aerial images.  
These problems have been often approached by an {\em oriented and densely packed object detection} task~\cite{DOTA, VEDAI, drone}, which is new while well-grounded and have attracted much attention in the past decade~\cite{SRBBS, RRPN, RRCNN, textboxes++, azimi2018towards}.
 \begin{figure}[t!]
    \centering
    \includegraphics[width=0.77\linewidth]{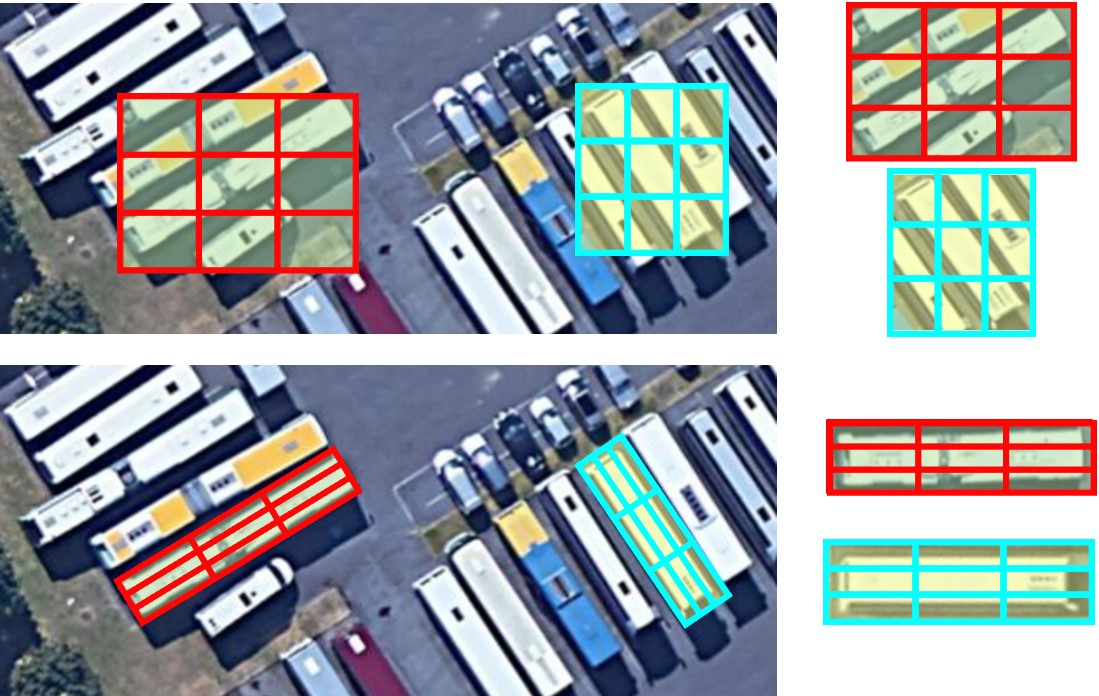}
    \vspace{1mm}
    \caption{{\bf {\em Horizontal} (top) {\em v.s.} {\em Rotated} RoI warping (bottom)} illustrated in an image with many densely packed objects.
    One horizontal RoI often contains several instances,  which leads ambiguity to the subsequent classification and location task. By contrast, a rotated RoI warping usually provides more accurate regions for instances and enables to better extract discriminative features for object detection.
    }
    \label{fig:RoIvsRRoI}
\end{figure}

Many of recent progresses on object detection in aerial images have benefited a lot from the R-CNN frameworks~\cite{r-cnn,Fast_RCNN,FasterR-CNN,VHR,long2017accurate,xiao2017airport,deng2017toward,drone,li2017object}.
These methods have reported promising detection performances, by using horizontal bounding boxes as {\em region of interests} (RoIs) and then relying on region-based features for category identification~\cite{VHR, long2017accurate, deng2017toward}. 
However, as observed in~\cite{DOTA, HRSC2016}, these {\em horizontal RoIs} (HROIs) typically lead to misalignments between the bounding boxes and objects. 
For instance, as shown in Fig.~\ref{fig:RoIvsRRoI}, due to the oriented and densely-distributed properties of objects in aerial images, several object instances are often crowded and contained by one HRoI. 
As a result, it usually turns to be difficult to train a detector for extracting object features and identifying the object's accurate localization. 

Instead of using horizontal bounding boxes, oriented bounding boxes have been alternatively employed to eliminate the mismatching between RRoIs and corresponding objects~\cite{DOTA,DLR3KMunichVehicle,HRSC2016}. 
In order to achieve high recalls at the phase of RRoI generation, a large number of anchors are required with different angles, scales and aspect ratios.
These methods have demonstrated promising potentials on detecting sparsely distributed objects~\cite{RRCNN, RSD, SRBBS, RRPN}. 
However, due to the highly diverse directions of objects in aerial images, it is often intractable to acquire accurate RRoIs to pair with all the objects in an aerial image by using RRoIs with limited directions.
Consequently, the elaborate design of RRoIs with as many directions and scales as possible usually suffers from its high computational complexity at region classification and localization phases. 

As the regular operations in conventional networks for object detection~\cite{Fast_RCNN} have limited generalization to rotation and scale variations, it is required of some orientation and scale-invariant in the design of RoIs and corresponding extracted features. To this end, Spatial Transformer~\cite{stn} and deformable convolution and RoI pooling~\cite{Deformable}  layers have been proposed to model the geometry variations. 
However, they are mainly designed for the general geometric deformation without using the oriented bounding box annotation. In the field of aerial images, there is only rigid deformation, and oriented bounding box annotation is available.
Thus, it is natural to argue that it is important to {\em extract rotation-invariant region features} and to {\em eliminate the misalignment between region features and objects} especially for densely packed ones. 

In this paper, we propose a module called RoI Transformer, targeting to achieve detection of oriented and densely-packed objects, by supervised RRoI learning and feature extraction based on position sensitive alignment through a two-stage framework~\cite{r-cnn,Fast_RCNN,FasterR-CNN,R-FCN,MaskRCNN}. 
It consists of two parts. The first is the {\em RRoI Learner}, which learns the transformation from HRoIs to RRoIs. The second is the {\em Rotated Position Sensitive RoI Align}, which extract the rotation-invariant feature extraction from the RRoI for subsequent objects classification and location regression.
To further improve the efficiency, we adopt a light head structure for all RoI-wise operations. 
We extensively test and evaluate the proposed RoI Transformer on two public datasets for object detection in aerial images \ie DOTA~\cite{DOTA} and HRSC2016~\cite{HRSC2016}, and compare it with state-of-the-art approaches, such as deformable PS RoI pooling~\cite{Deformable}. 
In summary, our contributions are in three-fold:
\begin{itemize}
    \item We propose a supervised rotated RoI leaner, which is a learnable module that can transform Horizontal RoIs to RRoIs. This design can not only effectively alleviate the misalignment between RoIs and objects, but also avoid a large amount of RRoIs designed for oriented object detection. 
    \item We designe a Rotated Position Sensitive RoI Alignment module for spatially invariant feature extraction, which can effectively boost the object classification and location regression. The module is a crucial design when using light-head RoI-wise operation, which grantees the efficiency and low complexity. 
    \item We achieve state-of-the-art performance on several public large-scale datasets for oriented object detection in aerial images. Experiments also show that the proposed RoI Transformer can be easily embedded into other detector architectures with significant detection performance improvements.
\end{itemize}
\section{Related Work}
\subsection{Oriented Bounding Box Regression}
Detecting oriented objects is an extension of general horizontal object detection. 
The objective of this problem is to locate and classify an object with orientation information, which is mainly tackled with methods based on region proposals. 
The HRoI based methods~\cite{R2CNN, DOTA} usually use a normal RoI Warping to extract feature from a HRoI, and regress position offsets relative to the ground truths. The HRoI based method exists a problem of misalignment between region feature and instance. The RRoI based methods~\cite{RRPN, RRCNN} usually use a Rotated RoI Warping to extract feature from a RRoI, and regress position offsets relative to the RRoI, which can avoid the problem of misalignment in a certain. 

However, the RRoI based method involves generating a lot of rotated proposals. The ~\cite{RRCNN} adopted the method in ~\cite{SRBBS} for rotated proposals. The SRBBS~\cite{SRBBS} is difficult to be embedded in the neural network, which would cost extra time for rotated proposal generation. The ~\cite{RRPN, RSD, R-DFPN, azimi2018towards} used a design of rotated anchor in RPN~\cite{FasterR-CNN}. However, the design is still time-consuming due to the dramatic increase in the number of anchors ($num\_scales\times num\_aspect\_ratios \times num\_angles$). For example, $3\times5\times6=90$ anchors at a location. A large amount of anchors increases the computation of parameters in the network, while also degrades the efficiency of matching between proposals and ground truths at the same time. Furthermore, directly matching between oriented bounding boxes (OBBs) is harder than that between horizontal bounding boxes(HBBs) because of the existence of plenty of redundant rotated anchors. Therefore, in the design of rotated anchors, both the ~\cite{RRPN, drbox} used a relaxed matching strategy. There are some anchors that do not achieve an IoU above 0.5 with any ground truth, but they are assigned to be True Positive samples, which can still cause the problem of misalignment.
In this work, we still use the horizontal anchors. The difference is that when the HRoIs are generated, we transform them into RRoIs by a light fully connected layer. Based on this strategy, it is unnecessary to increase the number of anchors. And a lot of precisely RRoIs can be acquired, which will boost the matching process. So we directly use the IoU between OBBs as a matching criterion, which can effectively avoid the problem of misalignment.


\subsection{Spatial-invariant Feature Extraction}
CNN frameworks have good properties for the generalization of translation-invariant features while showing poor performance on rotation and scale variations. For image feature extraction, the Spatial Transformer~\cite{stn} and deformable convolution~\cite{Deformable} are proposed for the modeling of arbitrary deformation. They are learned from the target tasks without extra supervision. 
For region feature extraction, the deformable RoI  pooling~\cite{Deformable} is proposed, which is achieved by offset learning for sampling grid of RoI pooling. It can better model the deformation at instance level compared to regular RoI warping~\cite{Fast_RCNN, MaskRCNN, R-FCN}. The STN and deformable modules are widely used for recognition in the field of scene text and aerial images~\cite{inceptext, deformable_faster, text-two-dimension, shi2016robust,xu2017deformable}. As for object detection in aerial images, there are more rotation and scale variations, but hardly nonrigid deformation. Therefore, our RoI Transformer only models the rigid spatial transformation, which is learned in the format of $(d_x, d_y, d_w, d_h, d_\theta)$. 
However, different from deformable RoI pooling, our RoI Transformer learns the offset with the supervision of ground truth. And the RRoIs can also be used for further rotated bounding box regression, which can also contribute to the object localization performance.

\subsection{Light RoI-wise Operations}
RoI-wise operation is the bottleneck of efficiency on two-stage algorithms because the computation are not shared. The Light-head R-CNN~\cite{light_head} is proposed to address this problem by using a larger separable convolution to get a thin feature. It also employs the PS RoI pooling~\cite{R-FCN} to further reduce the dimensionality of feature maps. A single fully connected layer is applied on the pooled features with the dimensionality of 10, which can significantly improve the speed of two-stage algorithms. In aerial images, there exist scenes where the number of instances is large. For example, over 800 instances are densely packed on a single $1024 \times 1024$ image. Our approach is similar to Deformable RoI pooling~\cite{Deformable} where the RoI-wise operations are conducted twice. The light-head design is also employed for efficiency guarantee.

 \begin{figure}[t!]
    \centering
    \includegraphics[width=0.7\linewidth]{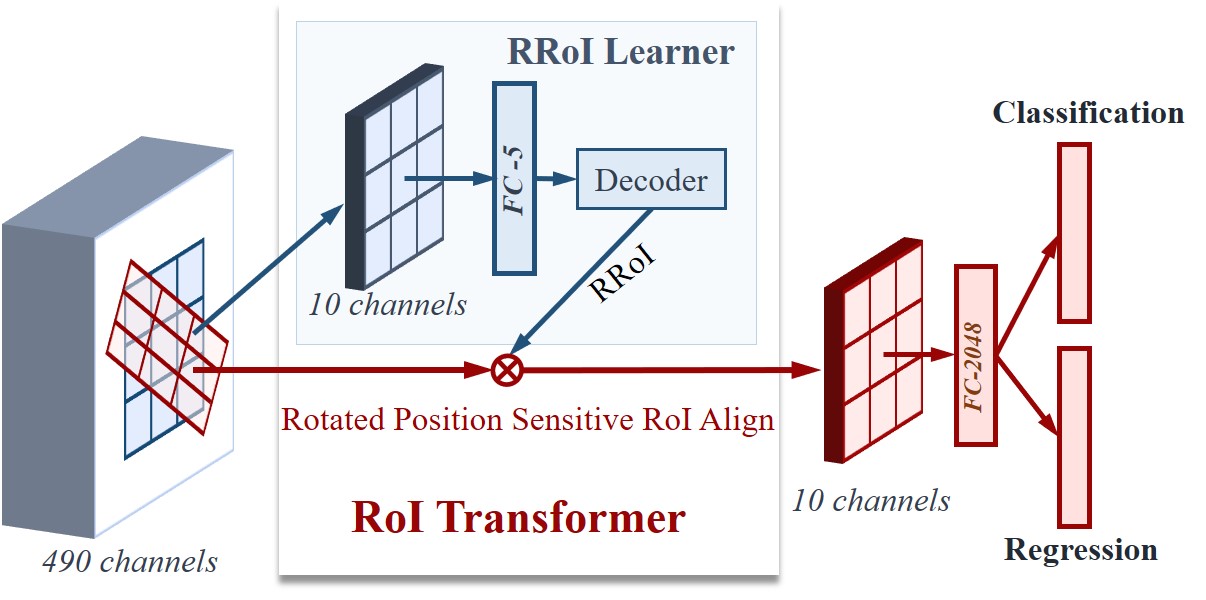}
    \caption{{\bf The architecture of RoI Transformer.} For each HRoI, it is passed to a RRoI learner. The RRoI learner in our network is a PS RoI Align followed by a fully connected layer with the dimension of 5 which regresses the offsets of RGT relative to HRoI. The Box decoder is at the end of RRoI Learner, which takes the HRoI and the offsets as input and outputs the decoded RRoIs. Then the feature map and the RRoI are passed to the RRoI warping for geometry robust feature extraction. The combination of RRoI Learner and RRoI warping form a RoI Transformer (RT). The geometry robust pooled feature from the RoI Transformer is then used for classification and RRoI regression.
    }
    \label{fig:cascadepipeline}
    \vspace{1mm}
\end{figure}
\begin{figure}[t!]
    \centering
    \includegraphics[width=0.77\linewidth]{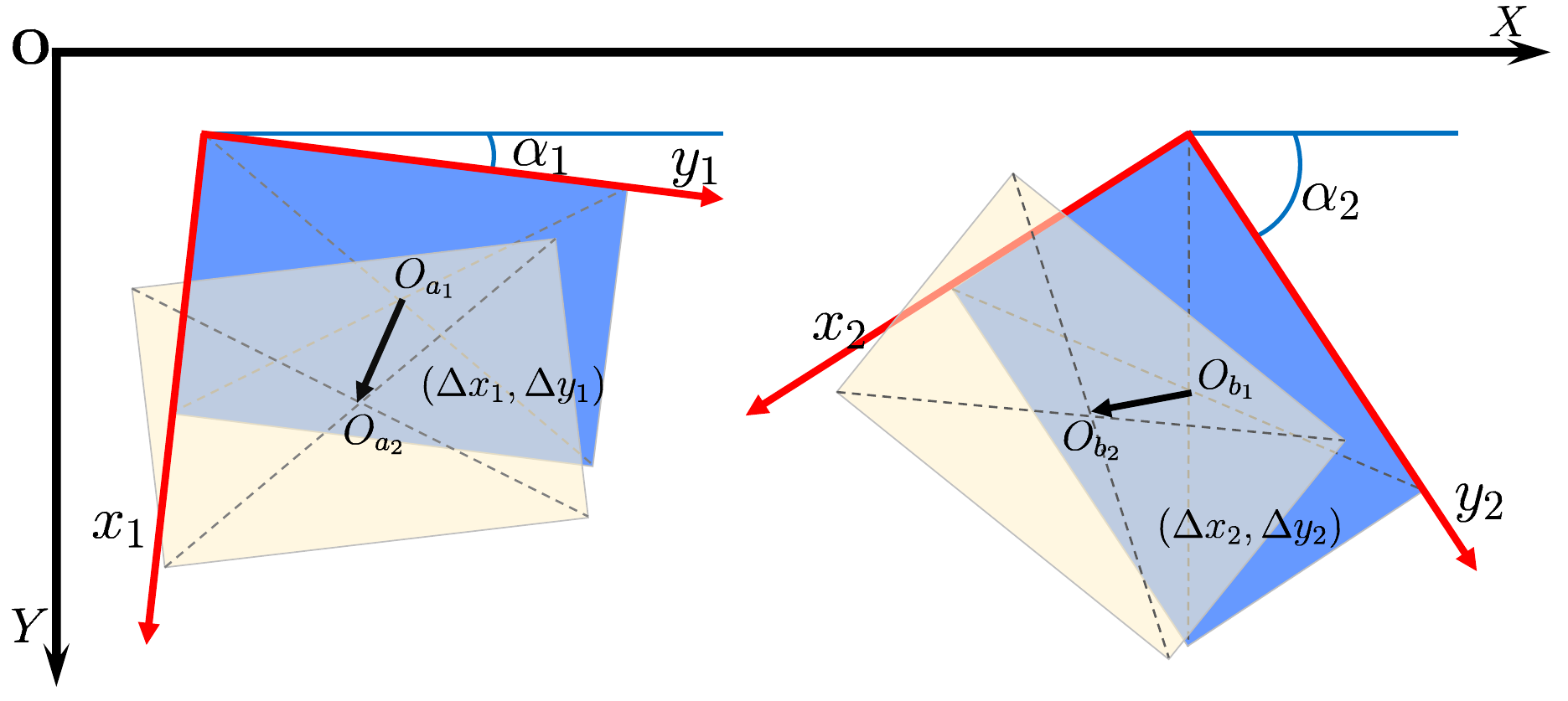}
    \caption{{\bf An example explaining the relative offset.} There are three coordinate systems. The $XOY$ is bound to the image. The $x_1O_1y_1$ and $x_2O_2y_2$ are bound to two RRoIs (blue rectangle) respectively. The yellow rectangle represents the RGT. The right two rectangles are obtained from the left two rectangles by translation and rotation while keeping the relative position unchanged.
    The $(\Delta x_1, \Delta y_1)$ is not equal to $(\Delta X_2, \Delta y_2)$ if they are all in the $XOY$. They are the same if $(\Delta x_1, \Delta y_1)$ falls in $x_1O_1y_1$ and $(\Delta X_2, \Delta y_2)$ in $(x_2O_2y_2)$. The $\alpha_1$ and $\alpha_2$ denote the angles of two RRoIs respectively.
    }
    \label{fig:coordinates}
    \vspace{1mm}
\end{figure}

\section{RoI Transformer}
In this section, we present details of our proposed \emph{ROI transformer}, which contains a trainable fully connected layer termed as \emph{RRoI Learner} and a \emph{RRoI warping} layer for learning the rotated RoIs from the estimated horizontal RoIs and then warping the feature maps to maintain the rotation invariance of deep features. Both of these two layers are differentiable for the end-to-end training.
The architecture is shown in Fig.\ref{fig:cascadepipeline}.
\subsection{RRoI Learner}
The RRoI learner aims at learning rotated RoIs from the feature map of horizontal RoIs. 
Suppose we have obtained $n$ horizontal RoIs denoted by $\{\mathcal{H}_i\}$ with the format of $(x,y,w,h)$ for predicted $2$D locations, width and height of a HRoI, the corresponding feature maps can be denoted as $\{\mathcal{F}_i\}$ with the same index.
Since every HRoI is the external rectangle of a RRoI in ideal scenarios, we are trying to infer the geometry of RRoIs from every feature map $\mathcal{F}_i$ using the fully connected layers. We follow the offset learning for object detection to devise the regression target as 
\begin{equation} \label{bbox_encoding}
\begin{split}
  t_x^* &= \tfrac{1}{w_r} \big( (x^*-x_r) \cos\theta_r + (y^*-y_r)\sin\theta_r \big),    \\
  t_y^* &= \tfrac{1}{h_r} \big( (y^*-y_r)\cos\theta_r - (x^*-x_r) \sin\theta_r) \big),  \\
  t_w^* &= \log \tfrac{w^*}{w_r}, \quad t_h^* = \log \tfrac{h^*}{h_r}, \\
  t_\theta^* &= \tfrac{1}{2\pi} \big( (\theta^*-\theta_r) \mod 2\pi \big),
\end{split}
\end{equation}
where $(x_r,y_r,w_r,h_r,\theta_r)$ is a stacked vector for representing location, width, height and orientation of a RRoI, respectively. $(x^*,y^*,w^*,h^*,\theta^*)$ is the ground truth parameters of an oriented bounding box. The modular operation is used to adjust the angle offset target $t_{\theta}^*$ that falls in $[0,2\pi)$ for the convenience of computation. Indeed, the target for HRoI regression is a special case of Eq. \eqref{bbox_encoding} if $\theta^* = \tfrac{3\pi}{2}$. The relative offsets are illustrated  in Fig.~\ref{fig:coordinates} as explanation. Mathematically, the fully connected layer outputs a vector $(t_x,t_y,t_w,t_h,t_{\theta})$  for every feature map $\mathcal{F}_i$ by 
\begin{equation}
\bm{t} = \mathcal{G}(\mathcal{F};\Theta),
\end{equation}
where $\mathcal{G}$ represents the fully connected layer and $\Theta$ is the weight parameters of $\mathcal{G}$ and $\mathcal{F}$ is the feature map for every HRoI.

While training the layer $\mathcal{G}$, we are about to match the input HRoIs and the ground truth of oriented bounding boxes (OBBs). 
{For the consideration of computational efficiency , the matching is between the HRoIs and axis-aligned bounding boxes over original ground truth.} 
Once an HRoI is matched, we set the $t_{\theta}^*$ directly by the definition in Eq. \eqref{bbox_encoding}. The loss function for optimization is used as Smooth L1 loss~\cite{r-cnn}. 
For the predicted $\bm{t}$ in every forward pass, we decode it from offset to the parameters of RRoI. That is to say, our proposed RRoI learner can learn the parameters of RRoI from the HRoI feature map  $\mathcal{F}$.

\subsection{Rotated Position Sensitive RoI Align}
\begin{figure}[t!]
    \centering
    \includegraphics[width=0.87\linewidth]{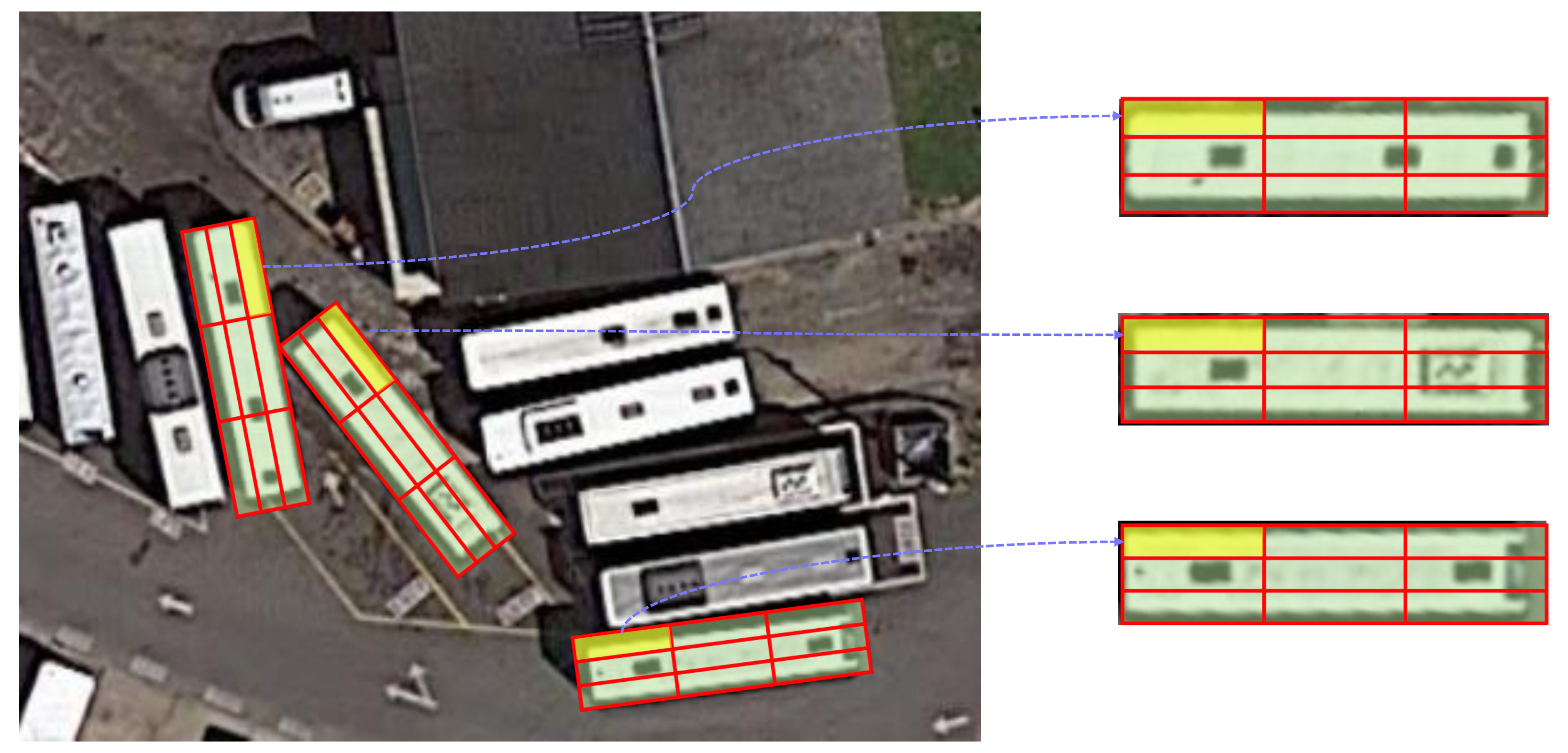}
    \caption{{\bf Rotated RoI warping} The shape of the warped feature is a horizontal rectangle (we use $3\times3$ for example here.) The sampling grid for RoI warping is determined by the RRoI $(x_r, y_r, w, h, \theta)$. We employ the image instead of feature map for better explanation. After RRoI warping, the extracted features are geometry robust. (The orientations of all the vehicles are the same).
    }
    \label{fig:R-RoI-warping}
\end{figure}
Once the parameters of RRoI are obtained, we are able to extract the rotation-invariant deep features for Oriented Object Detection.
Here, we propose the module of Rotated Position Sensitive (RPS) RoI Align to extract the rotation-invariant features within a network.

Given the input feature map $\mathcal{D}$ with $H \times W \times C$ channels and a RRoI $(x_r, y_r, w_r, h_r, \theta_r)$, where $(x_r, y_r)$ denotes the center of the RRoI and $(w_r, h_r)$ denotes the width and height of the RRoI. The $\theta_r$ gives the orientation of the RRoI.
The RPS RoI pooling divides the Rotated RoI into $K \times K$ bins and outputs a feature map $\mathcal{Y}$ with the shape of $(K\times K \times C)$. 
For the  bin with index $(i, j)$ $(0\leq i, j< K)$ of the output channel c$(0\leq c<C)$, we have
\begin{equation}\label{pooling}
    \mathcal{Y}_c(i,j)= \sum_{{}^{(x, y)\in bin(i,j)}} D_{i,j,c}(\mathcal{T}_\theta(x, y))/n_{ij},
\end{equation}
where the $D_{i,j,c}$ is a feature map out of the $K\times K \times C$ feature maps. The channel mapping is the same as the original Position Sensitive RoI pooling~\cite{R-FCN}. The $n_{ij}$ is the number of sampling locations in the bin. The $bin_{(i,j)}$
denotes the coordinates set $\{ i\frac{w_r}{k} + (s_x + 0.5)\frac{w_r}{k\times n}; s_x=0,1,... n-1 \}  \times 
\{ j\frac{h_r}{k} + (s_y + 0.5)\frac{h_r}{k\times n}; s_y=0,1,...n-1 \} $.
 And for each $(x, y)\in bin(i,j)$, it is converted to  $(x^{'},y^{'})$ by $\mathcal{T}_\theta$, where
\begin{equation}
\binom{x^{'}}{y^{'}}=\begin{pmatrix}
cos\theta & -sin\theta\\ 
sin\theta & cos\theta
\end{pmatrix} \binom{x-w_r/2}{y-h_r/2} + \binom{x_r}{y_r}
\end{equation}
Typically, Eq.~(\ref{pooling}) is implemented by bilinear interpolation.
\subsection{RoI Transformer for Oriented Object Detection}
The combination of RRoI Learner, and RPS RoI Align forms a RoI Transformer(RT) module. It can be used to replace the normal RoI warping operation.
The pooled feature from RT is rotation-invariant. And the RRoIs provide better initialization for later regression because the matched RRoI is closer to the RGT compared to the matched HRoI.
As mentioned before, a RRoI is a tuple with 5 elements ($x_r, y_r, w_r, h_r, \theta_r$). 
In order to eliminate ambiguity, we use $h$ to denote the short side and $w$ the long side of a RRoI. The orientation vertical to $h$  and falling in $[0, \pi]$ is chosen as the final direction of a RRoI. After all these operations, the ambiguity can be effectively avoided. And the operations are required to reduce the rotation variations. 

\begin{figure}[htp!]
    \centering
    \includegraphics[width=0.85\linewidth]{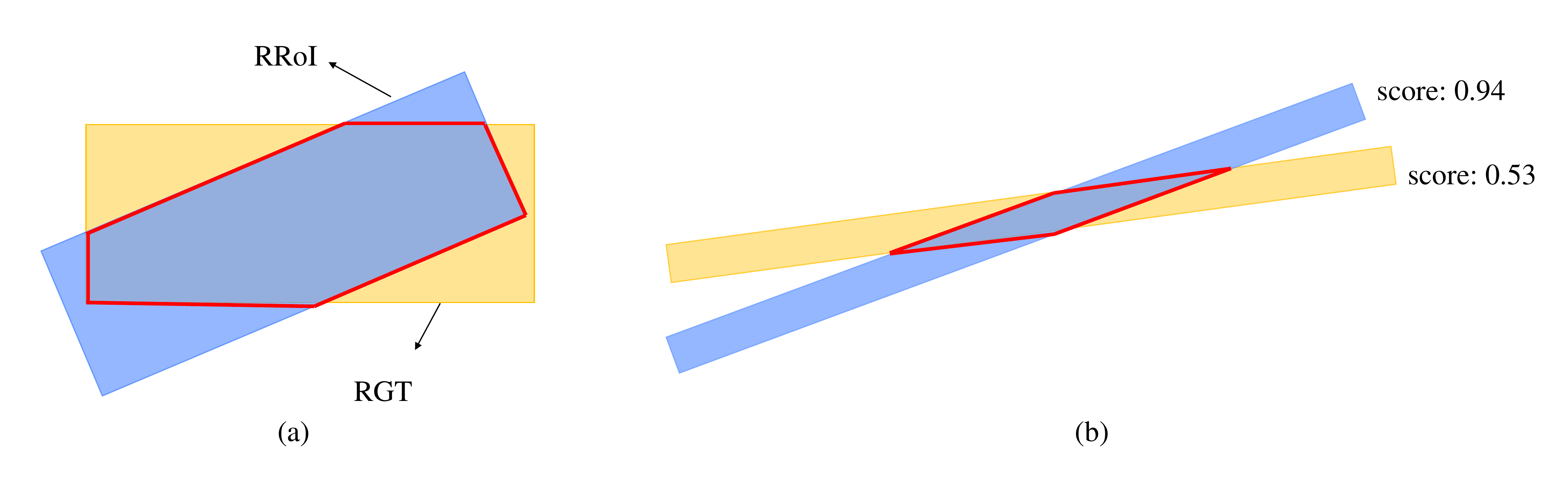}
    \caption{{\bf Examples of IoU between oriented bounding boxes(OBBs).} (a) IoU between a RRoI and a matched RGT. The red hexagon indicates the intersection area between RRoI and RGT. (b) The intersection between two long and thin bounding boxes. For long and thin bounding boxes, a slight jitter in the angle may lead to a very low IoU of the two boxes. The red quadrilateral is the intersection area. In such case, the predicted OBB with score of 0.53 can not be suppressed since the IoU is very low.
    }
    \label{fig:Vis_IoU}
\end{figure}

\paragraph{IoU between OBBs}\label{IoUBPolygon}
In common deep learning based detectors, there are two cases that IoU calculation is needed. The first lies in the matching process while the second is conducted for (Non-Maximum Suppression) NMS. The IoU between two OBBs can be calculated by Equation~\ref{IoU2}:
\begin{equation}\label{IoU2}
\mathit{IoU} = \frac{area(B_1\cap B_{2})}{area(B_1\cup B_{2})}
\end{equation}
where the $B_1$ and $B_2$ represent two OBBs, say, a RRoI and a RGT. The calculation of IoU between OBBs is similar with that between horizontal bounding boxes (HBBs). The only difference is that the IoU calculation for OBBs is performed within polygons as illustrated in Fig.~\ref{fig:Vis_IoU}. In our model, during the matching process, each RRoI is assigned to be True Positive if the IoU with any RGT is over 0.5. It is worth noting that although RRoI and RGT are both quadrilaterals, their intersection may be diverse polygons, e.g. a hexagon as shown in Fig~\ref{fig:Vis_IoU}(a).
For the long and thin bounding boxes, a slight jitter in the angle may cause the IoU of the two predicted OBBs to be very low, which would make the NMS difficult as can be seen in Fig.~\ref{fig:Vis_IoU}(b).

\paragraph{Targets Calculation}
After RRoI warping, the rotation-invariant feature can be acquired. Consistently, the offsets also need to be rotation-invariant. 
To achieve this goal, we use the relative offsets as explained in Fig.~\ref{fig:coordinates}. The main idea is to employ the coordinate system binding to the RRoI rather than the image for offsets calculation.
The Eq.~\eqref{bbox_encoding} is the derived formulation for relative offsets.



  


\section{Experiments and Analysis}
\subsection{Datasets}
For experiments, we choose two datasets, known as DOTA~\cite{DOTA} and HRSC2016~\cite{HRSC2016}, for oriented object detection in aerial images.

\begin{itemize}
    \item \label{shortname} \textbf{DOTA~\cite{DOTA}.} This is the largest dataset for object detection in aerial images with oriented bounding box annotations. It contains 2806 large size images. There are objects of 15 categories, including {\em Baseball diamond (BD), Ground track field (GTF), Small vehicle (SV),  Large vehicle (LV), Tennis court (TC), Basketball court (BC), Storage tank (ST), Soccer-ball field (SBF), Roundabout (RA), Swimming pool (SP), and  Helicopter (HC)}. The fully annotated DOTA images contain 188, 282 instances. The instances in this data set vary greatly in scale, orientation, and aspect ratio. As shown in ~\cite{DOTA}, the algorithms designed for regular horizontal object detection get modest performance on it.
    Like PASCAL VOC~\cite{PASCALVOC} and COCO~\cite{COCO}, the DOTA provides the evaluation server\footnote{\url{http://captain.whu.edu.cn/DOTAweb/}}. 
    
    We use both the training and validation sets for training, the testing set for test. We do a limited data augmentation. Specifically, we resize the image at two scales(1.0 and 0.5) for training and testing. After image rescaling, we crop a series of $1024 \times 1024$ patches from the original images with a stride of 824. For those categories with a small number of samples, we do a rotation augmentation randomly from 4 angles ($0, 90, 180, 270$) to simply avoid the effect of an imbalance between different categories. With all these processes, we obtain 37373 patches, which are much less than that in the official baseline implements (150, 342 patches)~\cite{DOTA}). 
    For testing experiments, the $1024 \times 1024$ patches are also employed. None of the other tricks is utilized except the stride for image sampling is set to 512. 
    \item \textbf{HRSC2016~\cite{HRSC2016}.} The HRSC2016~\cite{HRSC2016} is a challenging dataset for ship detection in aerial images. The images are collected from Google Earth. It contains 1061 images and more than 20 categories of ships in various appearances. The image size ranges from $300 \times 300$ to $1500 \times 900$. The training, validation and test set include 436 images, 181 images and 444 images, respectively. For data augmentation, we only adopt the horizontal flipping. And the images are resized to $(512, 800)$, where 512 represents the length of the short side and 800 the maximum length of an image.
\end{itemize}

\begin{table*}[t!]
\centering
\caption{Results of ablation studies. We used the {\em Light-Head R-CNN OBB} detector as our baseline. The leftmost column represents the optional settings for the RoI Transformer. In the right four experiments, we explored the appropriate setting for RoI Transformer.}
\vspace{2mm}
\begin{tabular}{c|c|cccc}
\hline
                       & Baseline & \multicolumn{4}{c}{Baseline + different settings}                                                             \\ \hline
RoI Transformer? &          & \checkmark & \checkmark & \checkmark & \checkmark \\
Light RRoI Learner?    &          &                           & \checkmark & \checkmark & \checkmark \\
Context region enlarge? &          &                           &                           & \checkmark & \checkmark \\
NMS on RRoIS?          &          & \checkmark & \checkmark & \checkmark &                           \\ \hline
mAP                    & 58.3     & 63.17                     & 63.39                     & 66.25                     & \bf{67.74}                     \\ \hline
\end{tabular}
\label{table:ablation}
\vspace{2mm}
\end{table*}

\begin{table*}[t!]
\centering
\caption{Comparisons with the state-of-the-art methods on HRSC2016.}
\vspace{2mm}
\setlength{\tabcolsep}{2.41mm}{
\begin{tabular}{c|cccccc|c}
\hline
method & CP~\cite{RRCNN}   & BL2~\cite{RRCNN}  & RC1~\cite{RRCNN}  & RC2~\cite{RRCNN}  & $R^{2}PN$~\cite{RSD}   & RRD~\cite{RRD}  & RoI Trans.    \\ \hline
mAP    & 55.7 & 69.6 & 75.7 & 75.7 & 79.6 & 84.3 &\textbf{ 86.2} \\ \hline
\end{tabular}
}
\end{table*}

\subsection{Implementation details}
    
\begin{table*}[t!]
\centering
\tiny
\caption{Comparisons with state-of-the-art detectors on DOTA~\cite{DOTA}. The short names for each category can be found in Section~\ref{shortname}. The FR-O indicates the {\em Faster R-CNN OBB} detector, which is the official baseline provided by DOTA~\cite{DOTA}. The RRPN indicates the {\em Rotation Region Proposal Networks}, which used a design of rotated anchor. The R2CNN means {\em Rotational Region CNN}, which is a HRoI-based method without using the RRoI warping operation. The RDFPN means the {\em Rotation Dense Feature Pyramid Netowrks}. It also used a design of Rotated anchors, and used a variation of FPN. The work in Yang et al.~\cite{yang2018position} is an extension of R-DFPN.}
\vspace{1.5mm}
\setlength{\tabcolsep}{1.1mm}{
\begin{tabular}{c|p{12pt}|ccccccccccccccc|c}
\hline
method & FPN & Plane & BD & Bridge & GTF & SV & LV & Ship & TC & BC & ST & SBF & RA & Harbor & SP & HC & mAP \\ 
\hline
FR-O~\cite{DOTA} & -- & 
79.42 & 77.13 & 17.7   & 64.05 & 35.3  & 38.02 & 37.16 & 89.41 & 69.64 & 59.28 & 50.3  & 52.91 & 47.89  & 47.4  & 
46.3 & 54.13 \\
RRPN~\cite{RRPN} & -- & 
80.94 & 65.75 & 35.34  & 67.44 & 59.92 & 50.91 & 55.81 & 90.67 & 66.92 & 72.39 & 55.06 & 52.23 & 55.14  & 53.35 & 48.22 & 61.01 \\
R2CNN~\cite{R2CNN} & -- & 
88.52 & 71.2  & 31.66  & 59.3  & 51.85 & 56.19 & 57.25 & 90.81 & 72.84 & 67.38 & 56.69 & 52.84 & 53.08  & 51.94 & 53.58 & 60.67 \\
R-DFPN~\cite{R-DFPN} & \checkmark & 
80.92 & 65.82 & 33.77  & 58.94 & 55.77 & 50.94 & 54.78 & 90.33 & 66.34 & 68.66 & 48.73 & 51.76 & 55.1   & 51.32 & 35.88 & 57.94 \\
Yang et al.~\cite{yang2018position} & \checkmark & 
81.25 & 71.41 & 36.53  & 67.44 & 61.16 & 50.91 & 56.6  & 90.67 & 68.09 & 72.39 & 55.06 & 55.6  & 62.44  & 53.35 & 51.47 & 62.29 \\ \hline
Baseline & -- & 
81.06 & 76.81 & 27.22  & 69.75 & 38.99 & 39.07 & 38.3  & 89.97 & 75.53 & 65.74 & {\bf 63.48} & 59.37 & 48.11  & 56.86 & 44.46 & 58.31 \\
DPSRP & -- & 
81.18 & 77.42 & 35.48  & 70.41 & 56.74 & 50.42 & 53.56 & 89.97 & {\bf 79.68} & 76.48 & 61.99 & 59.94 & 53.34  & {\bf 64.04} & 47.76 & 63.89 \\
RoI Transformer &-- & 
88.53 & 77.91 & 37.63  & 74.08 & 66.53 & 62.97 & 66.57 & 
90.5 & 79.46 & 76.75 & 59.04 & 56.73 & 62.54  & 61.29 & 
{\bf 55.56} & 67.74 \\
Baseline & \checkmark & 
88.02 & 76.99 & 36.7   & 72.54 & {\bf 70.15} & 61.79 & 75.77 & 90.14 & 73.81 & {\bf 85.04} & 56.57 & {\bf 62.63} & 53.3   & 59.54 & 41.91 & 66.95 \\
RoI Transformer & \checkmark & 
{\bf 88.64} & {\bf 78.52} & {\bf 43.44}  & {\bf 75.92} & 68.81 & {\bf 73.68} & {\bf 83.59} & {\bf 90.74} & 77.27 & 81.46 & 58.39 & 53.54 & {\bf 62.83}  & 58.93 & 47.67 & {\bf 69.56} \\ \hline
\end{tabular}
\vspace{2mm}
}\label{state-of-art}
\end{table*}
\begin{figure*}[t!]
    \centering
    \includegraphics[width=0.98\linewidth]{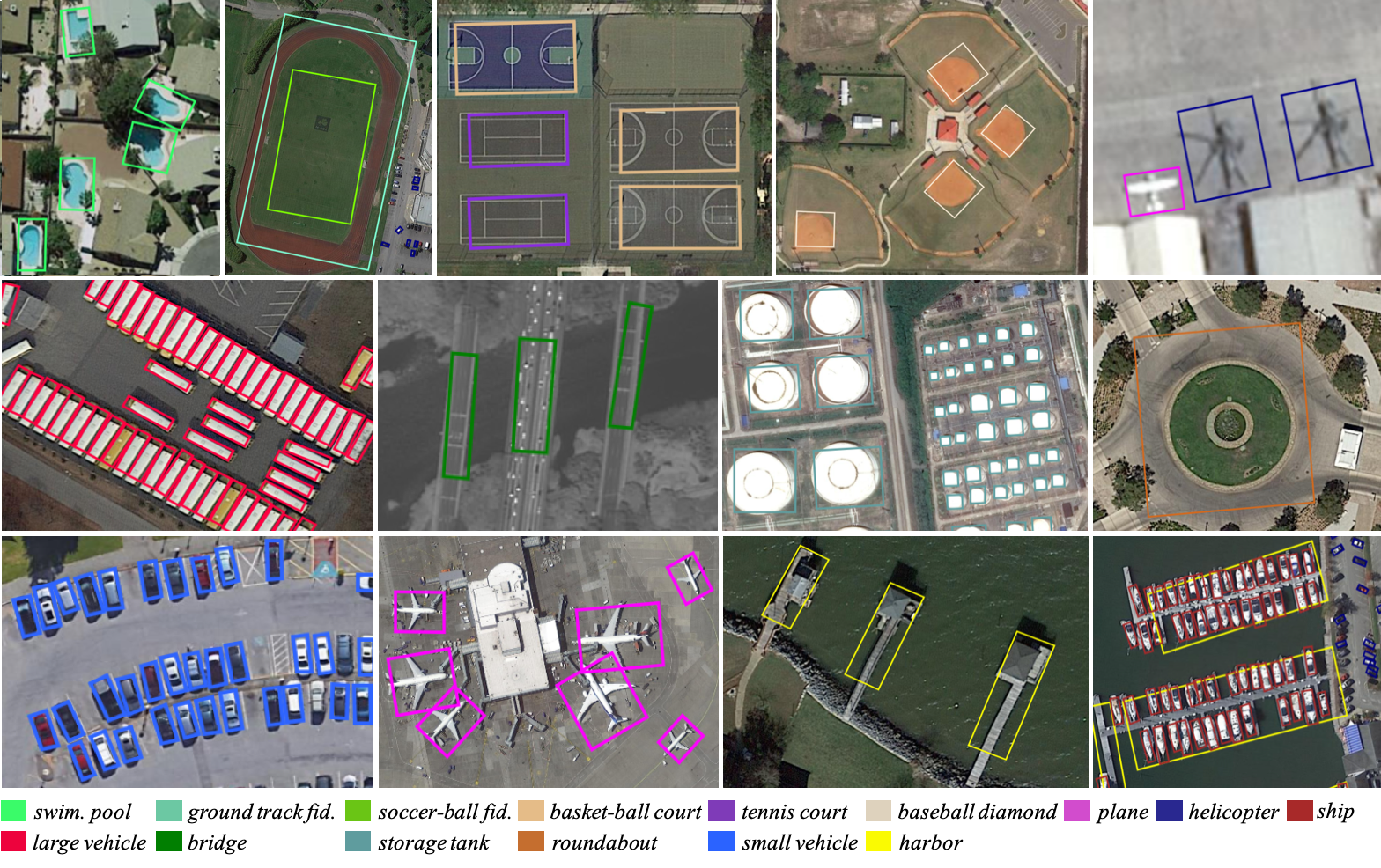}
    \caption{
    Visualization of detection results from RoI Transformer in DOTA.
    }
    \label{fig:vis_dota}
\end{figure*}

\begin{table}[t!]
\centering
\caption{Comparison of our RoI Transformer with deformable PS RoI pooling and Light-Head R-CNN OBB on accuracy, speed and memory. All the speed are tested on images with size of $1024\times 1024$ on a single TITAN X (Pascal). The time of post process (\ie NMS) was not included. The LR-O, DPSRP and RT denote the Light-Head R-CNN OBB, deformable Position Sensitive RoI pooling and RoI Transformer, respectively.}
\vspace{3mm}
\begin{tabular}{c|cccc}
\hline
method & mAP   & train speed & test speed & param   \\ \hline
LR-O   & 58.3  & \textbf{0.403 s}      & \textbf{0.141s}      &\textbf{ 273MB}   \\
DPSRP  & 63.89 & 0.445s       & 0.206s      & 273.2MB \\
RT     & \textbf{67.74} & 0.475s       & 0.17s       & \textbf{273MB}   \\ \hline
\end{tabular}
\vspace{-3mm}
\label{compare_with_deform}
\end{table}

\paragraph{Baseline Framework.} \label{baseline_frame}
For the experiments, we build the baseline network inspired from Light-Head R-CNN~\cite{light_head} with backbone ResNet101~\cite{resnet}. Our final detection performance is based on the FPN~\cite{FPN} network, while it is not employed in the ablation experiments for simplicity.

\begin{itemize}
    \item \textbf{Light-Head R-CNN OBB: }We modified the regression of fully-connected layer on the second stage to enable it to predict OBBs, similar to work in DOTA~\cite{DOTA}. The only difference is that we replace $({(x_i, y_i), i=1,2,3,4})$ with $(x, y, w, h, \theta)$ for the representation of an OBB.
    Since there is an additional param $\theta$, we do not double the regression loss as the original Light-Head R-CNN~\cite{light_head} does. 
    The hyperparameters of large separable convolutions we set is $k = 15, C mid = 256, C out = 490$. 
    And the OHEM~\cite{ohem} is not employed for sampling at the training phase. For RPN, we used 15 anchors same as original Light-Head R-CNN~\cite{light_head}. And the batch size of RPN~\cite{FasterR-CNN} is set to 512. Finally, there are 6000 RoIs from RPN before Non-maximum Suppression (NMS) and 800 RoIs after using NMS. Then 512 RoIs are sampled for the training of R-CNN.
    The learning rate is set to 0.0005 for the first 14 epochs and then divided by 10 for the last 4 epochs. 
    For testing, we adopt 6000 RoIs before NMS and 1000 after NMS processing.
    \item \textbf{Light-Head R-CNN OBB with FPN: }The Light-Head R-CNN OBB with FPN uses the FPN~\cite{FPN} as a backbone network. Since no source code was publicly available for Light-Head R-CNN based on FPN, our implementation details could be different. We simply added the large separable convolution on the feature of every level ${P_2, P_3, P_4, P_5}$. The hyperparameters of large separable convolution we set is $k = 15, Cmid = 64, Cout = 490$. 
    The batch size of RPN is set to be 512. There are 6000 RoIs from RPN before NMS and 600 RoIs after NMS processing. Then 512 RoIs are sampled for the training of R-CNN.
    The learning rate is set to 0.005 for the first 5 epochs and divided by a factor of 10 for the last 2 epochs.
\end{itemize}

\subsection{Comparison with Deformable PS RoI Pooling}
In order to validate that the performance is not from extra computation, we compared our performance with that of deformable PS RoI pooling, since both of them employed RoI warping operation to model the geometry variations. For experiments, we use the Light-Head R-CNN OBB as our baseline. The deformable PS RoI pooling and RoI Transformer are used to replace the PS RoI Align in the Light-Head R-CNN~\cite{light_head}.

\vspace{-2mm}
\paragraph{Complexity.} Both RoI Transformer and deformable RoI pooling have a light localisation network, which is a fully connected layer followed by the normal pooled feature. In our RoI Transformer, only 5 parameters($t_x, t_y, t_w, t_h, t_\theta$) are learned.
The deformable PS RoI pooling learns offsets for each bin, where the number of parameters is $7\times7\times2$ . So our module is designed lighter than deformable PS RoI pooling. As can be seen in Tab.~\ref{compare_with_deform}, our RoI Transformer model uses less memory (273MB compared to 273.2MB) and runs faster at the inference phase (0.17s compared to 0.206s per image). Because we use the light-head design, the memory savings are not obvious compared to deformable PS RoI pooling. 
However, RoI Transformer runs slower than deformable PS RoI pooling on training time (0.475s compared to 0.445s) since there is an extra matching process between the RRoIs and RGTs in training.

\begin{figure*}[t!]
    \centering
    \includegraphics[width=0.98\linewidth]{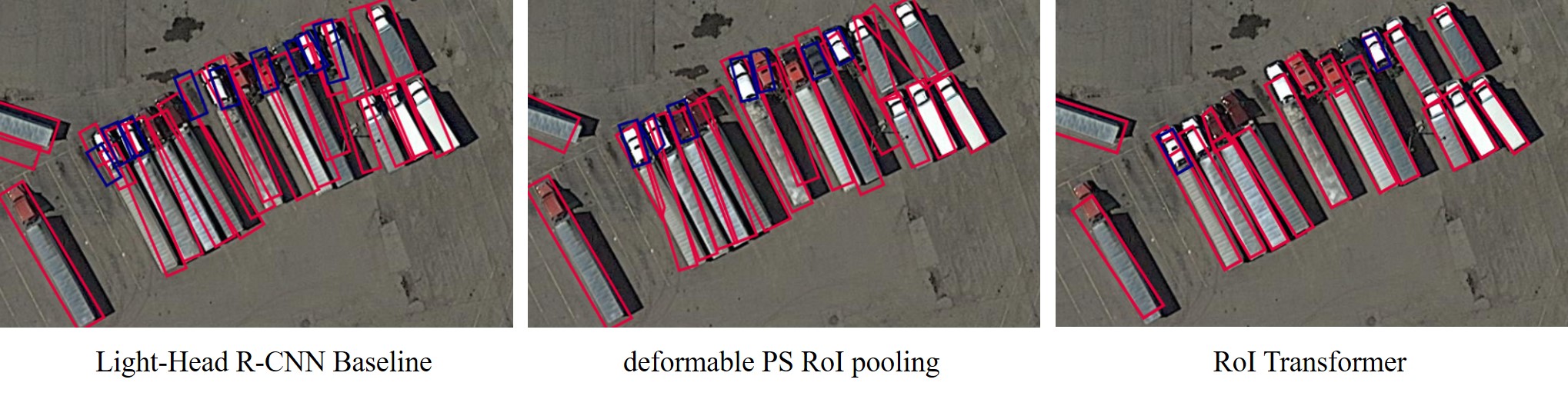}
    \caption{
    Visualization of detection on the scene where many densely packed instances exist. We select the predicted bounding boxes with scores above 0.1, and a NMS with threshold 0.1 is applied for duplicate removal. 
    }
    \vspace{2mm}
    \label{fig:dense_vis}
\end{figure*}

\begin{figure*}[t!]
    \centering
    \includegraphics[width=0.98\linewidth]{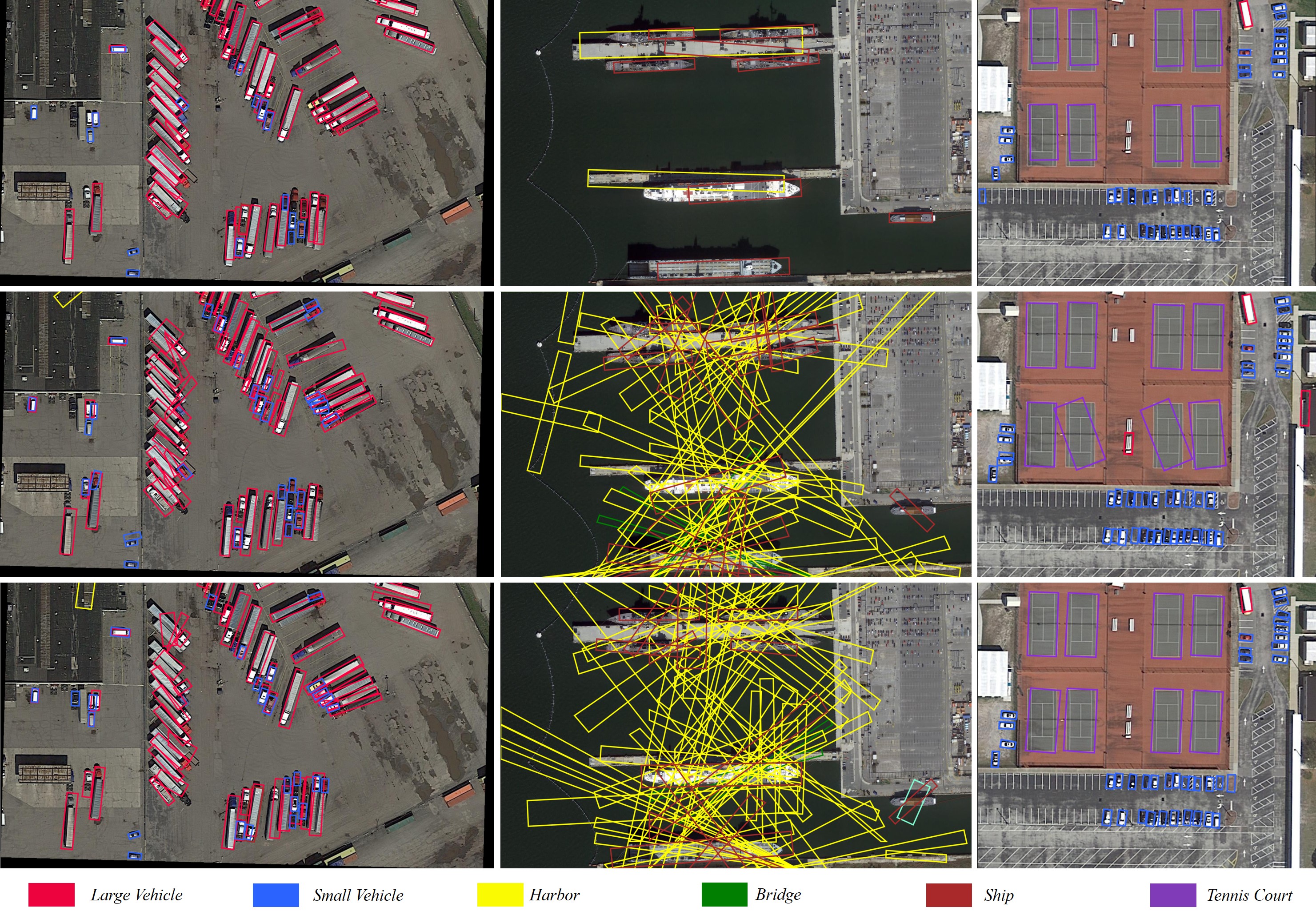}
    \caption{
    Visualization of detection results in DOTA. The first row shows the results from RoT Transformer. The second ros shows the results from Light-Head R-CNN OBB baseline. The last row shows the results from deformable PS RoI pooling. In the visualization, We select the predicted bounding boxes with scores above 0.1, and a NMS with threshold 0.1 is applied for duplicate removal.
    }
    \label{fig:vis_dota4}
\end{figure*}

\begin{figure*}[t!]
    \centering
    \includegraphics[width=0.98\linewidth]{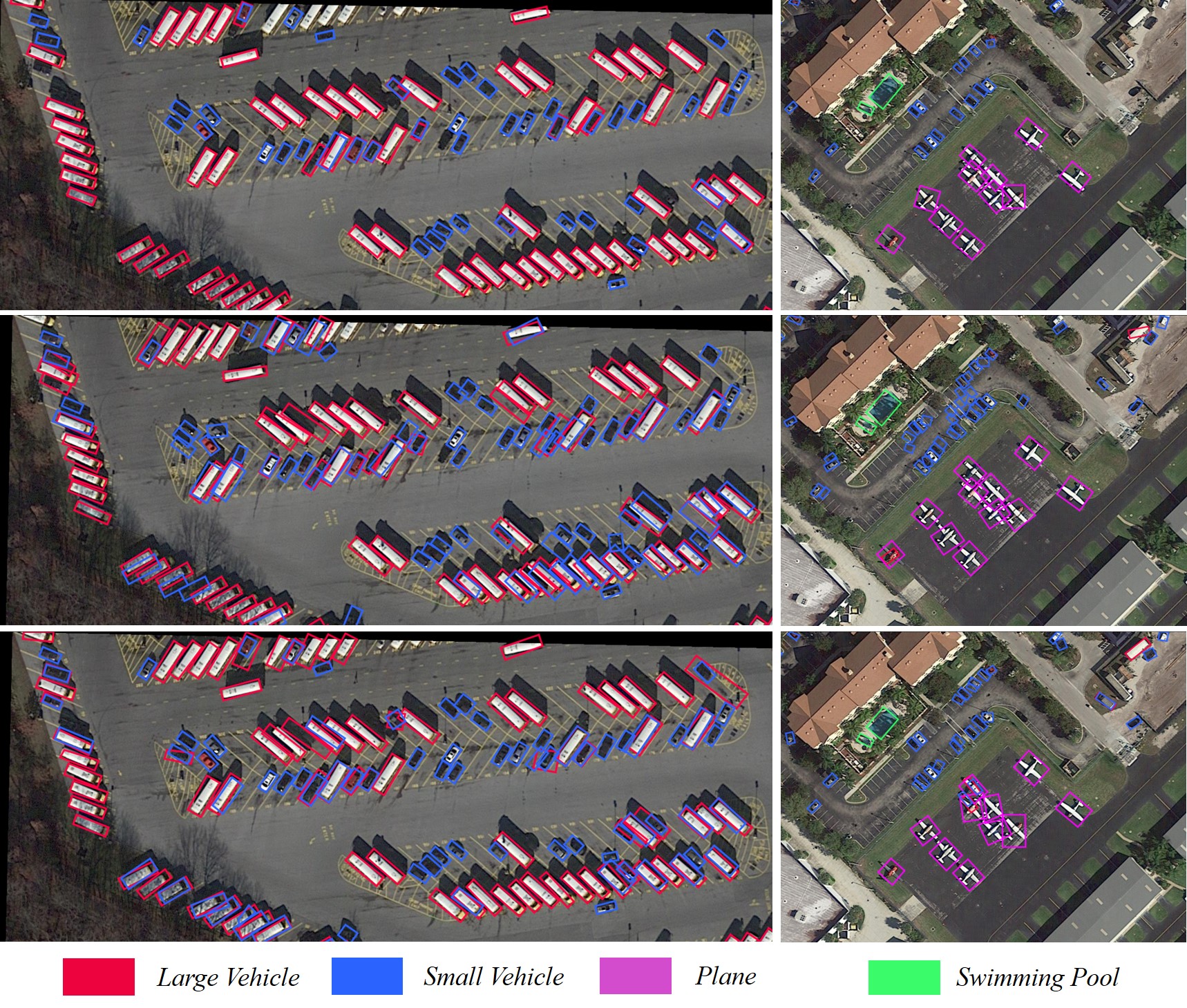}
    \caption{
    Visualization of detection results in DOTA. The first row shows the results from RoT Transformer. The second ros shows the results from Light-Head R-CNN OBB baseline. The last row shows the results from deformable PS RoI pooling. In the visualization, We select the predicted bounding boxes with scores above 0.1, and a NMS with threshold 0.1 is applied for duplicate removal.
    }
    \label{fig:vis_dota3}
\end{figure*}

\vspace{-2mm}
\paragraph{Detection Accuracy.}
The comparison results are shown in Tab.~\ref{compare_with_deform}. The deformable PS RoI pooling outperforms the Light-Head R-CNN OBB Baseline by 5.6 percents. While there is only 1.4 points improvement for R-FCN~\cite{R-FCN} on Pascal VOC~\cite{PASCALVOC} as pointed out in ~\cite{Deformable}. It shows that the geometry modeling is more important for object detection in aerial images. But the deformable PS RoI pooling is much lower than our RoI Transformer by 3.85 points.
We argue that there are two reasons: 1) Our RoI Transformer can better model the geometry variations in aerial images. 2)
The regression targets of deformable PS RoI pooling are still relative to the HRoI rather than using the boundary of the offsets. Our regression targets are relative to the RRoI, which gives a better initialization for regression. The visualization of some detection results based on Light-Head R-CNN OBB Baseline, Deformable Position Sensitive RoI pooling and RoI Transformer are shown in Fig.~\ref{fig:dense_vis}, Fig.~\ref{fig:vis_dota4} and Fig.~\ref{fig:vis_dota3}, respectively. The results in Fig.~\ref{fig:dense_vis} and the first column of Fig.~\ref{fig:vis_dota4} are taken from the same large image. It shows that RoI Transformer can precisely locate the instances in scenes with densely packed ones. And the Light-Head R-CNN OBB baseline and the deformable RoI pooling show worse accuracy performance on the  localization of instances. It is worth noting that the head of truck is misclassified to be small vehicle (the blue bounding box) for the three methods as shown in Fig.~\ref{fig:dense_vis} and Fig.~\ref{fig:vis_dota4}. While our proposed RoI Transformer has the least number of misclassified instances. The second column in Fig~\ref{fig:vis_dota4} is a complex scene containing long and thin instances, where both Light-Head R-CNN OBB baseline and deformable PS RoI pooling generate many False Negatives. And these False Negatives are hard to be suppressed by NMS due to the reason as explained in Fig.~\ref{fig:Vis_IoU}(b). Benefiting from the consistency between region feature and instance, the detection results based on RoI Transformer generate much fewer False Negatives.


\subsection{Ablation Studies}
We conduct a serial of ablation experiments on DOTA to analyze the accuracy of our proposed RoI Transformer. We use the Light-Head R-CNN OBB as our baseline. Then gradually change the settings. When simply add the RoI Transformer, there is a 4.87 point improvement in mAP. The other settings are discussed in the following.

\paragraph{Light RRoI Learner.} In order to guarantee the efficiency, we directly apply a fully connected layer with output dimension of 5 on the pooled features from the HRoI warping. As a comparison, we also tried more fully connected layers for the RRoI learner, as shown at the first and second columns in Tab.~\ref{table:ablation}. We find there is little drop (0.22 point) on mAP when we add on more fully connected layer with output dimension of 2048 for the RRoI leaner. The little accuracy degradation should be due to the fact that the additional fully connected layer with higher dimensionality requires a longer time for convergence.

\begin{figure}[t!]
    \centering
    \includegraphics[width=0.7\linewidth]{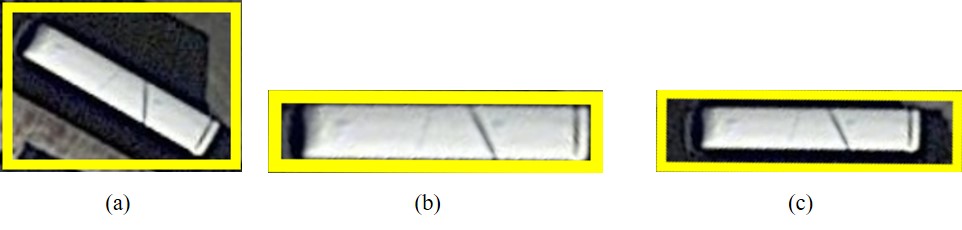}
    \vspace{-2mm}
    \caption{Comparison of 3 kinds of region for feature extraction.
 (a) The Horizontal Region. (b) The rectified Region after RRoI Warping. (c) The rectified Region with appropriate context after RRoI warping. 
    }
    \label{fig:ContextRegion}
\end{figure}

\paragraph{Contextual RRoI.}
As pointed in ~\cite{tinyfaces, RRPN}, appropriate enlargement of the RoI will promote the performance. A horizontal RoI may contain much background while a precisely RRoI hardly contains redundant background as explained in the Fig.~\ref{fig:ContextRegion}. Complete abandon of contextual information will make it difficult to classify and locate the instance even for the human. Therefore, it is necessary to enlarge the region of the feature with an appropriate degree. Here, we enlarge the long side of RRoI by a factor of 1.2 and the short side by 1.4. The enlargement of RRoI improves AP by 2.86 points, as shown in Tab.~\ref{table:ablation}

\begin{figure*}[ht!]
    \centering
    \includegraphics[width=0.95\linewidth]{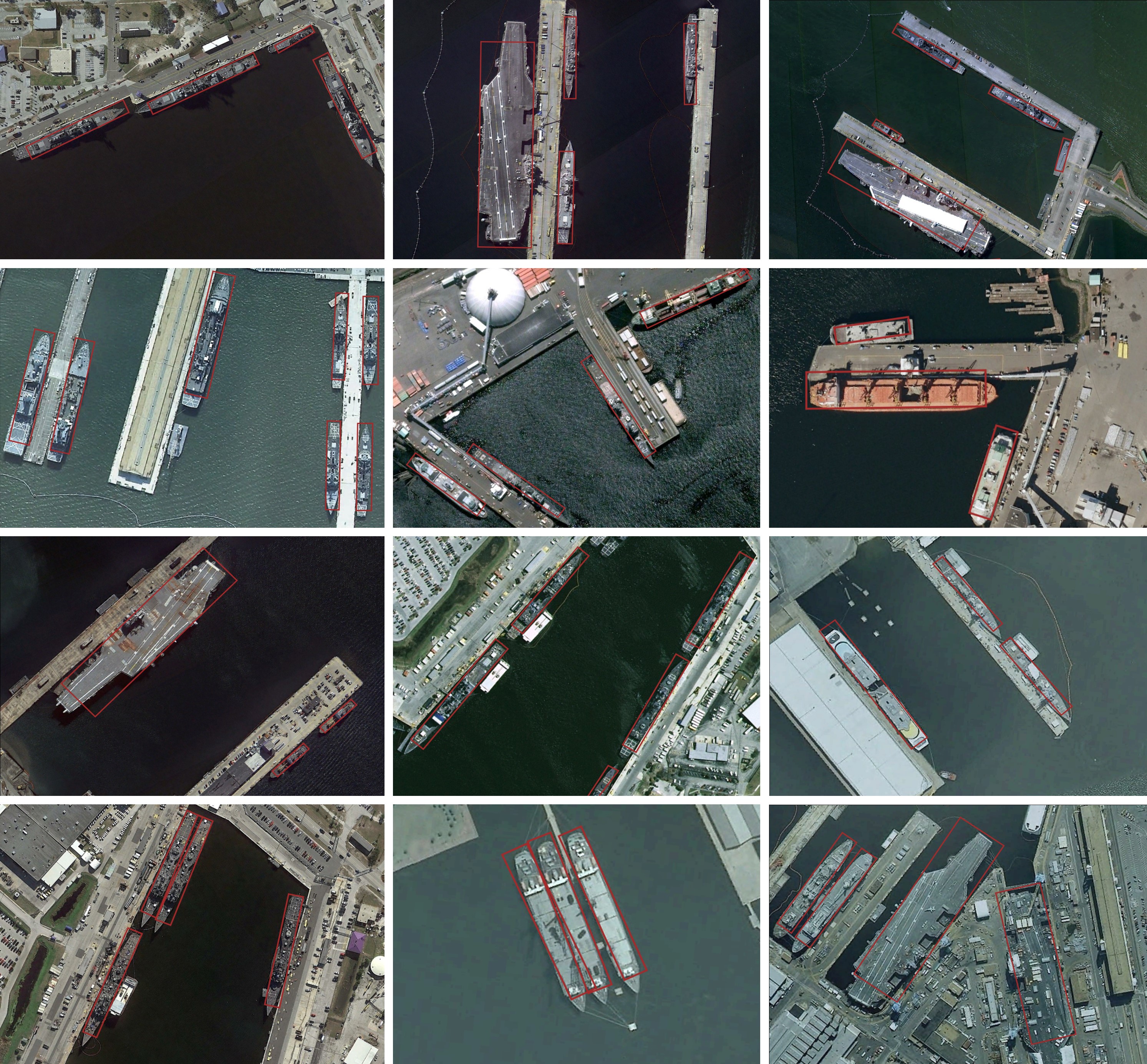}
    \caption{{\bf Visualization of detection results from RoI Transformer in HRSC2016}. We select the predicted bounding boxes with scores above 0.1, and a NMS with threshold 0.1 is applied for duplicate removal.
    }
    \label{fig:vis_HRSC}
\end{figure*}

\paragraph{NMS on RRoIs.}Since the obtained RoIs are rotated, there is flexibility for us to decide whether to conduct another NMS on the RRoIs transformed from the HRoIs. This comparison is shown in the last two columns of Tab.~\ref{table:ablation}. We find there is $\sim$ 1.5 points improvement in mAP if we remove the NMS. This is reasonable because there are more RoIs without additional NMS, which could increase the recall.

\subsection{Comparisons with the State-of-the-art}
We compared the performance of our proposed RoI Transformer with the state-of-the-art algorithms on two datasets DOTA~\cite{DOTA} and HRSC2016~\cite{HRSC2016}. The settings are described in Sec.~\ref{baseline_frame}, and we just replace the Position Sensitive RoI Align with our proposed RoI Transformer. Our baseline and RoI Transformer results are obtained without using ohem~\cite{ohem} at the training phase.

\paragraph{Results on DOTA.} We compared our results with the state-of-the-arts in DOTA. Note the RRPN~\cite{RRPN} and R2CNN~\cite{R2CNN} are originally used for text scene detection. The results are a re-implemented version for DOTA by a third-party\footnote{\url{https://github.com/DetectionTeamUCAS/RRPN_Faster-RCNN_Tensorflow}}. As can be seen in Tab.~\ref{state-of-art}, our RoI Transformer achieved the mAP of 67.74 for DOTA , it outperforms the previous the state-of-the-art without FPN (61.01) by 6.71 points. And it even outperforms the previous FPN based method by 5.45 points. With FPN, the Light-Head OBB Baseline achieved mAP of 66.95, which outperforms the previous state-of-the-art detectors, but still slightly lower than RoI Transformer. When RoI Transformer is added on Light-Head OBB FPN Baseline, it gets improvement by 2.6 points in mAP reaching the peak at 69.56. This indicates that the proposed RoI Transformer can be easily embedded in other frameworks and significantly improve the detection performance. Besides, there is a significant improvement in densely packed small instances. (e.g. the small vehicles, large vehicles, and ships). For example, the detection performance for the ship category gains an improvement of 26.34 points compared to the previous best result (57.25) achieved by R2CNN~\cite{R2CNN}. Some qualitative results of RoI Transformer on DOTA are given in Fig~\ref{fig:vis_dota}. 

\paragraph{Results on HRSC2016.} The HRSC2016 contains a lot of thin and long ship instances with arbitrary orientation. We use 4 scales $\{ 64^{2}, 128^{2}, 256^{2}, 512^{2} \}$ and 5 aspect ratios $\{ 1/3, 1/2, 1, 2, 3 \}$, yielding $k = 20$ anchors for RPN initialization. This is because there is more aspect ratio variations in HRSC, but relatively fewer scale changes. The other settings are the same as those in ~\ref{baseline_frame}.  We conduct the experiments without FPN which still achieves the best performance on mAP. Specifically, based on our proposed method, the mAP can reach 86.16, 1.86 higher than that of RRD~\cite{RRD}. Note that the RRD is designed using  SSD~\cite{SSD} for oriented object detection, which utilizes multi-layers for feature extraction with 13 different aspect ratios of boxes$\{1, 2, 3, 5, 7, 9, 15, 1/2, 1/3, 1/5, 1/7, 1/9, 1/15 \}$. While our proposed framework just employs the final output features with only 5 aspect ratios of boxes. In Fig.~\ref{fig:vis_HRSC}, we visualize some detection results in HRSC2016. The orientation of the ship is evenly distributed over $2\pi$. In the last row, there are closely arranged ships, which are difficult to distinguish by horizontal rectangles. While our proposed RoI Transformer can handle the above mentioned problems effectively. The detected  incomplete ship in the third picture of the last row proves the strong stability of our proposed RoI Transformer detection method.


\section{Conclusion}
In this paper, we proposed a module called RoI Transformer to model the geometry transformation and solve the problem of misalignment between region feature and objects. The design brings significant improvements for oriented object detection on the challenging DOTA and HRSC with negligible computation cost increase. While the deformable module is a well-designed structure to model the geometry transformation, which is widely used for oriented object detection. The comprehensive comparisons with deformable RoI pooling solidly verified that our model is more reasonable when oriented bounding box annotations are available. So, it can be inferred that our module can be an optional substitution of deformable RoI pooling for oriented object detection.

{\small
\bibliographystyle{ieeetr}
\bibliography{egbib}
}

\end{document}